%% file: acl2021.tex
\documentclass[11pt,a4paper]{article}
\usepackage[hyperref]{acl2021}
\usepackage{times}
\usepackage{latexsym}
\usepackage{multirow}

\usepackage{graphicx}
\usepackage{amsmath}

\usepackage{amssymb}
\usepackage[normalem]{ulem}
\usepackage{amsthm}
\usepackage{color}
\usepackage{xspace}
\usepackage{subfigure}
\usepackage{booktabs}
\usepackage{array,tabularx}
\usepackage[ruled,vlined]{algorithm2e}
\usepackage{algpseudocode}
\usepackage{amsfonts}

\usepackage{xspace}
\usepackage{microtype}
\usepackage[ruled,vlined]{algorithm2e}
\usepackage{algpseudocode}
\usepackage{microtype}

\aclfinalcopy 
\def\aclpaperid{4273} 

\title{Could you give me a hint? \\Generating inference graphs for defeasible reasoning}

\author{Aman Madaan~\thanks{\hspace{0.5em}Equal Contribution}\hspace{0.5em}, Dheeraj Rajagopal~\footnotemark[1]\hspace{0.5em}, Niket Tandon~\footnotemark[1]\hspace{0.5em}$^\dagger$,  Yiming Yang, Eduard Hovy \\
  Language Technologies Institute, Carnegie Mellon University, Pittsburgh, PA, USA \\
  $^\dagger$ Allen Institute for Artificial Intelligence, Seattle, WA, USA \\ 
  \texttt{\{dheeraj,amadaan,yiming,hovy\}@cs.cmu.edu} \\ \texttt{\{nikett\}@allenai.org} \\} 

\date{}
\input{sections/custom_commands}
\begin{document}
\maketitle

\input{sections/abstract}
\input{sections/introduction}


\input{sections/rq1}
\input{sections/rq2}

\input{sections/conclusion}

\bibliographystyle{acl_natbib}
\bibliography{acl2021}
\newpage
\clearpage

\appendix
\input{sections/appendix}


\end{document}

%% file: sections/custom_commands.tex
\definecolor{Red}{rgb}{1,0,0}
\definecolor{Green}{rgb}{0,1,0}
\definecolor{Blue}{rgb}{0,0,1}
\definecolor{Red}{rgb}{0.9,0,0}
\definecolor{Orange}{rgb}{1,0.5,0}
\definecolor{yellow}{rgb}{0.65,0.6,0}
\definecolor{cadmiumgreen}{rgb}{0.2, 0.7, 0.24}
\newcommand{\todo}[1]{\textcolor{Red}{[#1 \textsc{--TODO}]}}

\newcommand{\secref}[1]{\S\ref{#1}}

\newcommand{\V}[1]{\mathbf{#1}}

\newcommand{\X}[1]{\mathbf{\mathrm{S}}}
\newcommand{\Z}[1]{\mathbf{\mathrm{C^-}}}
\newcommand{\VV}[1]{\mathbf{\mathrm{C^+}}}
\newcommand{\W}[1]{\mathbf{\mathrm{M^-}}}
\newcommand{\U}[1]{\mathbf{\mathrm{S^-}}}
\newcommand{\Y}[1]{\mathbf{\mathrm{M^+}}}
\newcommand{\LL}[1]{\mathbf{\mathrm{H^-}}}
\newcommand{\M}[1]{\mathbf{\mathrm{H^+}}}

\newcommand{\red}[1]{\textcolor{Red}{#1}}

\newcommand{\cadmiumgreen}[1]{\textcolor{cadmiumgreen}{#1}}

\newcommand{\wiqa}{\textsc{wiqa}\xspace}

\newcommand{\dotf}{\textsc{dot}\xspace}

\newcommand{\lm}{$\mathcal{L}$\xspace}

\newcommand{\tf}{\textsc{t5}\xspace}

\newcommand{\bleu}{\textsc{BLEU}\xspace}
\newcommand{\rouge}{\textsc{ROUGE}\xspace}
\newcommand{\upd}{$\mathbf{U}$\xspace}
\newcommand{\hypo}{$\mathbf{H}$\xspace}
\newcommand{\pre}{$\mathbf{P}$\xspace}

\newcommand{\atomic}{\textsc{atomic}\xspace}
\newcommand{\snli}{\textsc{snli}\xspace}
\newcommand{\social}{\textsc{social}\xspace}

\newcommand{\sts}{sequence-to-sequence\xspace}
\newcommand{\rqone}{\textsc{rq1}\xspace}
\newcommand{\rqtwo}{\textsc{rq2}\xspace}

\newcommand{\inten}{\textit{Intensifies}\xspace}
\newcommand{\atten}{\textit{Attenuates}\xspace}
\newcommand{\dques}{(\pre, \hypo, \upd)\xspace}

\newcommand{\squishlist}{
  \begin{list}{$\bullet$}
    { \setlength{\itemsep}{0pt}      \setlength{\parsep}{3pt}
      \setlength{\topsep}{3pt}       \setlength{\partopsep}{0pt}
      \setlength{\leftmargin}{1.5em} \setlength{\labelwidth}{1em}
      \setlength{\labelsep}{0.5em} } }
\newcommand{\reallysquishlist}{
  \begin{list}{$\bullet$}
    { \setlength{\itemsep}{0pt}    \setlength{\parsep}{0pt}
      \setlength{\topsep}{0pt}     \setlength{\partopsep}{0pt}
      \setlength{\leftmargin}{0.2em} \setlength{\labelwidth}{0.2em}
      \setlength{\labelsep}{0.2em} } }

 \newcommand{\squishend}{
     \end{list} 
 }
 

%% file: sections/abstract.tex
\begin{abstract}
Defeasible reasoning is a mode of reasoning where conclusions can be overturned by taking into account new evidence. A commonly used method in cognitive science and logic literature is to handcraft argumentation supporting inference graphs. While humans find inference graphs very useful for reasoning, constructing them at scale is difficult. In this paper, we automatically generate such inference graphs through transfer learning from a related NLP task that shares the kind of reasoning that inference graphs support. Through automated metrics and human evaluation, we find that our method generates meaningful graphs for the defeasible inference task. Human accuracy on this task improves by 20\% by consulting the generated graphs.
Our findings open up exciting new research avenues for cases where machine reasoning can help human reasoning.\footnote{A dataset of 230,000 influence graphs for each defeasible query is located at: \url{https://tinyurl.com/defeasiblegraphs}.}
\end{abstract}

%% file: sections/introduction.tex
\section{Introduction}
Defeasible inference~\cite{rudinger-etal-2020-thinking} is a mode of reasoning in which given a premise \pre (Rob went for a hike), a hypothesis \hypo (Rob saw an elephant, it was pink) may be weakened or overturned in light of new evidence i.e., an update \upd (Rob often has hallucinations).
Given the non-monotonic nature of this reasoning, humans find it challenging to master this task~\cite{Morgan2004TheNO}.
This problem has been widely studied in classical AI through logic \cite{Israel1980WhatsWW,McCarthy1981SOMEPP}, and in cognitive science through argumentative models \cite{Pollock1987DefeasibleR}. A prominent approach is to support defeasible inference through argumentations by constructing an \emph{inference graph} \citep{Pollock2009ARS}.

\begin{figure*}[!ht]
\centering
{\includegraphics[scale=0.50]{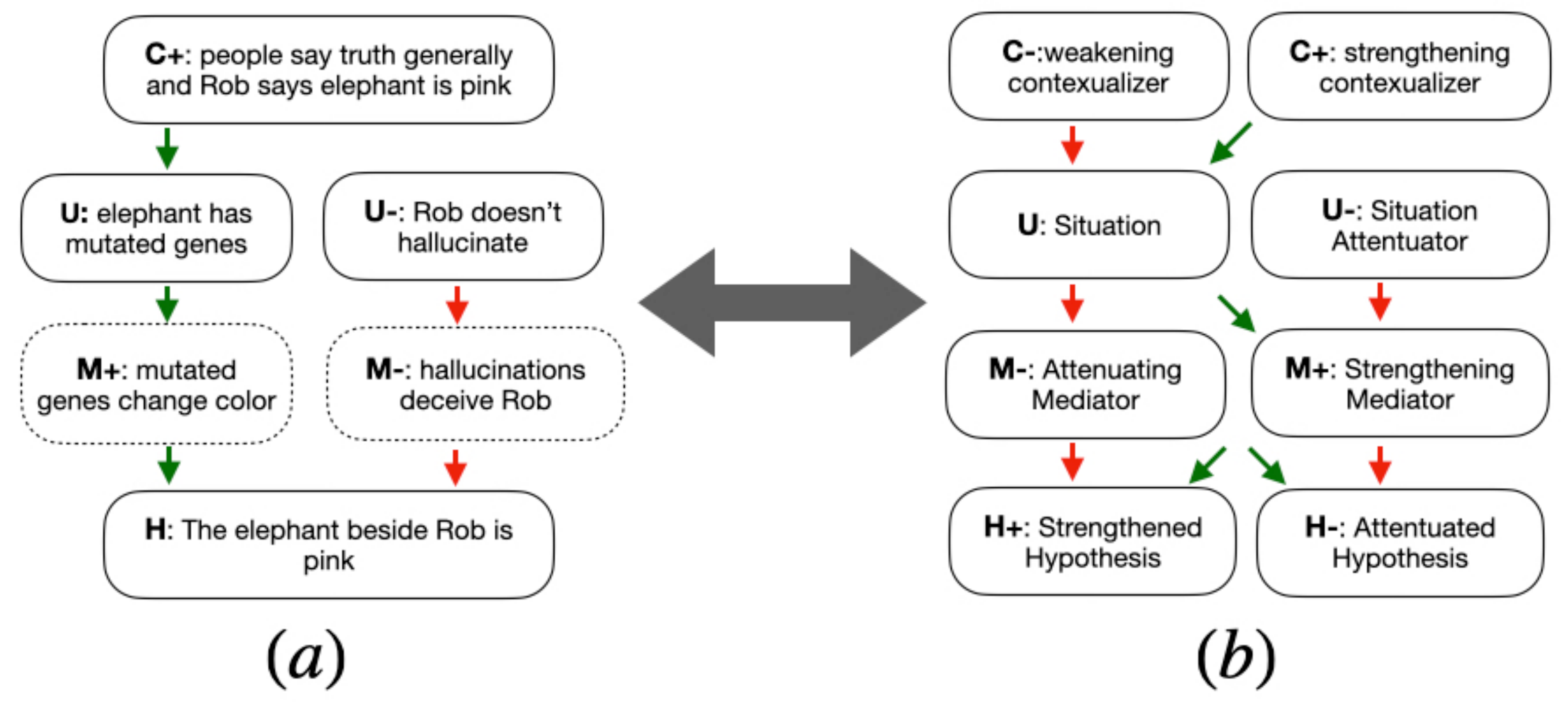}}
\caption{(\textbf{a}) An example of an Inference Graph adapted from \citet{Pollock2009ARS} and (\textbf{b}) Structure of an Influence Graph adapted from \wiqa \citep{tandon2019wiqa} dataset. The adapted influence graph incorporates the contextualizers, mediators, hypotheses and situations, making them useful for defeasible reasoning.}
\label{fig:igraph}
\end{figure*}

Despite their prominence \cite{Bentahar2010ATO}, argumentative models are not scalable because an inference graph needs to be handcrafted for every example.
Recently, \citet{rudinger-etal-2020-thinking} proposed two auxiliary tasks related to defeasible inference: (i) an NLI task to predict whether an update \upd would weaken or strengthen a hypothesis \hypo, and (ii) a generative task to generate an update \upd given a premise \pre and a hypothesis \hypo. 
However, this only addresses a part of the problem because their inference is still not supported by the line of reasoning that a human typically uses to solve this task, namely mediators (e.g., hallucinations can be deceptive) and contextualizers (some elephants can have mutated gene which makes them look different) that are inherently embedded in an inference graph, limiting their utility for humans (figure \ref{fig:igraph}).

In this paper, we adopt the concept of an inference graph for defeasible reasoning from cognitive science and provide a computational model to make their generation scalable. 
Training such a model would require a large amount of annotated inference graphs, which will be too expensive to obtain.
Instead, our solution is to draw a parallel to a related reasoning task in NLP~\cite{tandon2019wiqa}, where the reasoning is supported by a graph that we find has similarities with the kind of reasoning that an inference graph supports.
We train a model that can learn from the NLP task and effectively transfer it to generate inference graphs. Such transfer learning is made possible due to the powerful seq-to-seq neural language models that did not exist before.

The contributions of this paper are the answers to the following two research questions:
\squishlist
    \item[\textbf{\rqone}] Can we automate the construction of the  argumentation supporting inference graphs? In \S\ref{sec:rq1}, we show that we can effectively construct meaningful graphs using transfer learning.
    \item[\textbf{\rqtwo}] Can our generated graphs help improve human performance? In \S\ref{sec:rq2}, we show that humans leverage generated graphs to improve their performance on a previously reported benchmark.
\squishend

%% file: sections/rq1.tex

\section{\rqone: Generating argumentation supporting Inference Graphs}
\label{sec:rq1}

We start by drawing parallels to a counterfactual reasoning task in NLP - the \wiqa \citep{tandon2019wiqa} task. 
\wiqa consists of a set of procedural passages, each accompanied by a human-curated \textit{influence graph}.
The influence graph captures the causal influences between the events in the context of the process described by the passage.
We draw a connection between inference graphs~\cite{Pollock2009ARS} and influence graphs~\cite{tandon2019wiqa} by drawing parallels between their reasoning structures. In essence, each inference graph from \citet{Pollock1987DefeasibleR} can be instantiated via an influence graph from \citet{tandon2019wiqa} by interpreting the nodes in both the graphs as follows~(Figure~\ref{fig:igraph}):


\begin{enumerate}
\item[i.] \textbf{Contextualizers (C):} these nodes set the context around a situation and connect to the \pre in some way.
\item[ii.] \textbf{Updates (U):} these nodes are new situations that emerge which might overturn an inference.
\item[iii.] \textbf{Hypothesis (H):} Hypothesis nodes describes the outcome/conclusion of the situation.
\item[iv.] \textbf{Mediators (M):} Mediators are nodes that help bridge the knowledge gap between a situation and a hypothesis node by explaining their connection explicitly. 
\end{enumerate}

Figure \ref{fig:igraph} presents an example to highlight the similarities between the two graphs by labeling an example node adapted from \cite{Pollock2009ARS}, and the structure of the influence graph from \cite{tandon2019wiqa} with the four node types that we defined above.
A~\cadmiumgreen{green} edge indicates that the source node has a positive influence on the target node, and a~\red{red} edge indicates a negative influence. 
Further, each node can either act as a \emph{strengthener} (\textbf{+}) or a \emph{weakener} (\textbf{-}) for the hypothesis.
Consequently, these graphs can support similar type of reasoning e.g., the effect of \upd on \hypo and how this can change in light of external influences (\textbf{C}) is captured by graph paths $\V{C}$\textbf{+} to \upd and from \upd via a mediator node ($\V{M}$\textbf{+}/$\V{M}$\textbf{-}) to \hypo.
Inspired by these similarities, we hypothesize that influence graphs can be used to supplement defeasible reasoning.


\subsection{Influence Graphs Generation}
\label{sec:igraphgeneration}

\begin{figure}[!h]
\centering
\includegraphics[scale=0.45]{
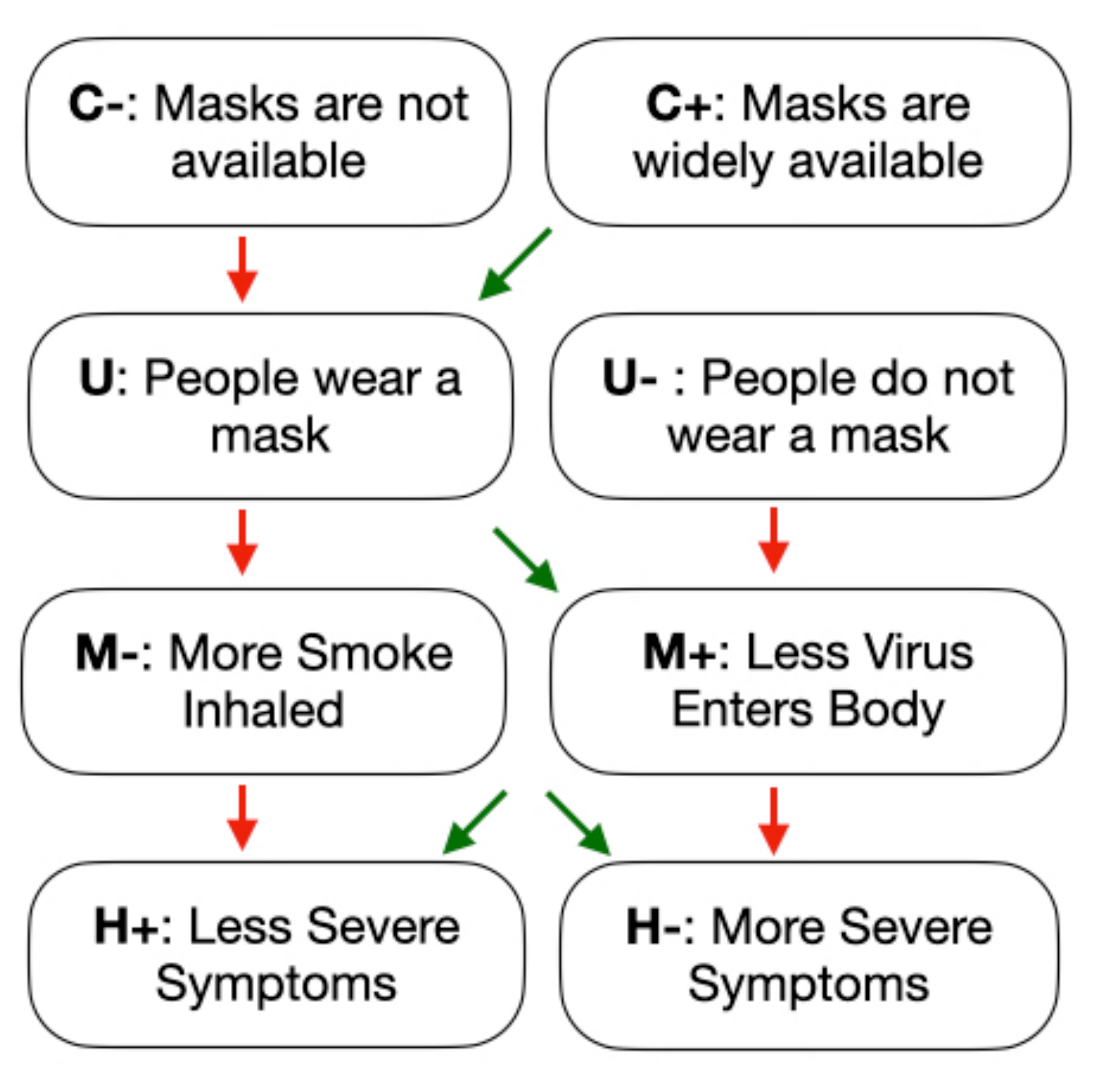}
\caption{An example of an influence graph similar to ones in WIQA that we train on.}
\label{fig:igraph_example_wiqa}
\end{figure}

To obtain an influence graph for each defeasible query, we perform a zero-shot transfer from \wiqa~\cite{tandon2019wiqa}, a corpus of 2100 (passage, influence graphs) pairs.\footnote{Dataset details in the Appendix~\secref{sec:datasetdetails}.}.
\paragraph{Training :}
We treat influence graph generation as a \sts mapping task.
We leverage \wiqa to derive parallel data $\{(\V{seq}_{ip}^i, \V{seq}_{op}^i)\}_{i=1}^{N}$ for the task.
Let $(\V{T}_i, \V{G}_i)$ be a sample in \wiqa, where $\V{T}_i$ is the passage text (e.g. describing how viruses spread), and $\V{G}_i$ is the corresponding influence graph~(e.g., Figure~\ref{fig:igraph_example_wiqa}). To create tokens of the input sequence $\V{seq}_{ip}^i$, the model trains best with explicit markers:\footnote{An example shown in Appendix~\secref{sec:sampleinputoutput}.} 

\setlength{\belowdisplayskip}{0pt} \setlength{\belowdisplayshortskip}{0pt}
\setlength{\abovedisplayskip}{0pt} \setlength{\abovedisplayshortskip}{0pt}

\begin{small}
\begin{align}
\label{eq:input-tokens}
\V{seq}_{ip}^i &= \text{Premise: } \V{T}_i \mid \text{Update: } \V{U}_i \mid \text{less/ more} \text{: } \V{H}_i    
\end{align}
\end{small}
 where $\V{T}_i$ is the passage text (e.g. steps describing how viruses spread) and $\V{U}_i$ and $\V{H}_i$ are nodes of $\V{G}_i$ (these are phrases as shown in Figure \ref{fig:igraph_example_wiqa}).

The output $\V{seq}_{op}^i$ is set to a \dotf-string representation of the corresponding influence graph $\V{G}_i$, as such a representation was shown to be effective at extracting high-quality graphs~\cite{madaan-yang-2021-neural} from free-form text using language models~(examples in the appendix).
Thus, each passage-graph pair $(\V{T}_i, \V{G}_i)$ from \wiqa is mapped to an input-output pair $\mathcal{D} = (\V{seq}_{ip}^i, \V{seq}_{op}^i)$.
We use this corpus to fine-tune an autoregressive language model \lm for graph generation.
Essentially, the fine-tuned \lm allows us to efficiently sample an influence graph for a given input sequence $\V{seq}_{ip}^j$ by drawing samples from $\V{G}_j \sim P_{\theta}(y \mid \V{seq}_{ip}^j)$ using greedy sampling, where $\theta$ denotes the parameters of the language model. 

\paragraph{Zero-shot Transfer to Defeasible Inference :}
We use the model $\mathcal{L}$ trained on \wiqa to generate inference graphs on the defeasible inference dataset by~\citet{rudinger-etal-2020-thinking}. 
We obtain an influence graph for each defeasible input \dques by converting it to an input sequence that can be fed to \lm by filling the template~(\ref{eq:input-tokens}).
This conversion from \dques to template~(\ref{eq:input-tokens}) is done by setting the premise \pre as the context passage $\V{T}$, the update \upd as the node $\V{U}$, and the attenuated and strengthened outcomes are simulated by prefixing the hypothesis \hypo with the tokens \textit{Less} and \textit{More} respectively. 
This input is then passed to the \lm to generate an influence graph.

\paragraph{Results on Influence Graph Generation}
We use T5-11B~\cite{raffel2020exploring} fine-tuned on $\mathcal{D}$ derived from \wiqa~(\secref{sec:igraphgeneration}) as our graph generation language model 
(\lm).
All the graphs generated by our model were in valid \dotf format.
We use the standard generation metrics \bleu~\cite{papineni2002bleu} and \rouge~\cite{lin2004rouge} to evaluate \lm on the test split of \wiqa.
Each node $N_i$ in the reference graph is compared with the corresponding generated node $\hat{N_i}$ using $\texttt{BLEU}(N_i, \hat{N_i})$ (Node-BLEU).
Further, node-edge-node pairs (neighbors) $(N_i, N_j)$ and $(\hat{N_i}, \hat{N_j})$ are compared using Rel-BLEU = $\texttt{HM}(\texttt{BLEU}(N_i, \hat{N_i}), \texttt{BLEU}(N_j, \hat{N_j}))$
where $\texttt{HM}$ is the harmonic mean.
These metrics are averaged over the graph (i.e., across the nodes and the edges), and further averaged across the corpus.
We perform these experiments across two different language models: \textsc{gpt-2-medium}~\cite{radford2019language} and \textsc{t5-11b}.
Finally, we calculate the overlap in the edge structures of the reference and generated graphs match as Edge-\textsc{match}\%.
We report the numbers in Table~\ref{tab:auto-gen-quality}, and include a random baseline for reference.
A random baseline will correctly generate the nodes $\V{S}$, $\V{H}$+, and $\V{H}$- as they are part of the query~($\frac{3}{8}$ nodes).
As neither of these nodes are connected to another, the random baseline will likely not generate any node pair correctly~( Rel-\bleu $\sim$ 0).
Since two unique graph structures are possible~\cite{tandon2019wiqa}, a random baseline would get Edge-match $\sim$ 50\%.
Table~\ref{tab:auto-gen-quality} shows that our \tf-based model is able to generate syntactically valid (high edge-match) and semantically meaningful graphs.
Additionally, we find that our generated graphs are helpful to humans on a downstream task, as described next.

\begin{table}[ht]
\centering
\small
\begin{tabular}{@{}lrrr@{}}
\toprule
Model  & Random & \textsc{gpt-2-medium} & \textsc{t5-11b}  \\ \midrule
Node-\bleu  & 37.5 & 46.05 & \textbf{50.94}\\
Rel-\bleu   & 0.0 & 19.34 & \textbf{33.01}   \\
Edge-match\%& 50.0 & 92.86 & \textbf{97.63} \\
\bottomrule
\end{tabular}
\caption{Results on automated metrics showing that our \textsc{t5-11b} model is able to generate very accurate graph structure and meaningful nodes that sufficiently match the reference nodes.}
\label{tab:auto-gen-quality}
\end{table}


%% file: sections/rq2.tex
\section{\rqtwo: Do generated graphs help humans at defeasible reasoning?}
\label{sec:rq2}

\paragraph{Human Evaluation}
\citet{rudinger-etal-2020-thinking} performed a human evaluation on 2000 defeasible queries, where given \dques, the task was to label the nature of the effect of \upd on \hypo as \inten or \atten.
Three human judges labeled each query, and the majority label was then compared with the ground-truth to ascertain the accuracy.
In their setup, human judges were collectively right on 1745 samples (correct pool) and wrong on 255 samples (wrong pool).
We create a challenging pool of 510 queries for the human judges by combining the 255 queries in the wrong pool with 255 queries sampled from the correct pool, giving a baseline accuracy of 50\% for this eval pool.
Each query in this pool is supplemented with a generated influence graph~(\secref{sec:rq1}).\footnote{Discussion on IRB exemption in Section~\secref{sec:irbexemption}.}
We found that our generated influence graphs showed high-levels of redundancy in contextualizers and mediators, with about 46\% of the generated influence graphs repeating these nodes.
We found that humans find it simpler to follow positive chains of influence, so to reduce their cognitive load, we post-process each influence graph to only retain the strengthening contextualizer~(Figure~\ref{fig:igraph}), the situation (\upd), the strengthening mediator (\textbf{M+}), and the hypothesis (\hypo).

In order to establish comparable gains, we replicate the evaluation setup of \citet{rudinger-etal-2020-thinking}~by using use the same Amazon Mechanical Turk template and the instruction set, and the same pool of 230 qualified annotators that \citet{rudinger-etal-2020-thinking} selected based on a paid qualification test, in which the workers were asked to answer SNLI queries of varying levels of difficulty. We paid slightly above \$15 per hour for the tasks. 

For each query, in addition to answering the defeasible question, three judges were asked to evaluate the augmented influence graphs on two aspects:

\begin{enumerate}
\item [i)] \textbf{Is the influence graph useful?} The judges were allowed to select from the following: 
\begin{enumerate}
\item \textit{helpful}: the graph was crucial in helping towards answering the question
\item \textit{relevant but not helpful}: the graph had the right topic (relevant to the question) but did not help in answering the question.
\item \textit{irrelevant or misleading}: the graph was irrelevant to the question or misled the human judge to a wrong answer. 
\end{enumerate}


\item [ii)] \textbf{Why is the influence graph useful?} The judges were given an option to highlight the most useful aspect of the generated influence graph. They were allowed to tag one or more of the following aspects as the most helpful: i) Extraneous node, ii) Mediating node, and iii) Structure of the graph.
\end{enumerate}
We summarize the key findings below.

\paragraph{Finding 1: influence graphs are helpful and relevant}
As Table~\ref{tab:human-eval-relevance} shows, a large majority of the human judges found the influence graphs to be helpful or relevant.
We calculate the inter-annotator agreement for this question using majority-agreement
$ = \frac{1}{N}\sum_{i=1}^{N} \texttt{ma}_i$
where $\texttt{ma}_i$ indicates a majority agreement for the $i^{th}$ sample (i.e., at least 2 out of 3 judges agreed on the label for the sample).
The majority-agreement (\texttt{ma}) on these labels was 0.83.
The judges marked about 25\% of the graphs as relevant but not helpful. The graphs in such cases were on topic but not helpful in answering the query, thereby distinguishing the cases when the graph was crucial in reaching the correct answer. Finally, we note that the graphs provided as hints could have been helpful in two ways: by helping the human annotators arrive at the answer, or by reinforcing their mental picture that helped them in making the right decision. Future research in this direction is needed to study these aspects in depth.

\begin{table}[ht]
\centering
\begin{tabular}{ll}
\toprule
Helpful  & 47.25 \\
Relevant but not helpful & 25.09 \\
Irrelevant or misleading   & 10.58 \\
No majority agreement & 17.05 \\ \bottomrule
\end{tabular}
\caption{Helpfulness of the augmentations.}
\label{tab:human-eval-relevance}
\end{table}

\paragraph{Finding 2: Mediators are the most helpful for defeasible queries}

For every sample, we asked the human judges to mark which parts of the graph was the most helpful (as shown in Figure \ref{appendix:mturk} in Appendix~\secref{sec:samplehiit}). The judges could select more than one aspect of the graph if they found multiple useful aspects. 
Table~\ref{tab:ig-aspects} shows the percentage of human judges that selected the particular graph aspect as most helpful. We observe that 49.48\% of the judges who found the graphs useful indicated the mediator node as the most helpful.
This indicates that while there may be other events that impact \upd and \hypo, the mediating events are the most informative in determining the type of link between them. 

\begin{table}[ht]
\centering
\begin{tabular}{lr}
\toprule
Aspect & \% marked useful \\ \midrule
Mediator & 49.48 \\
Extraneous                      & 32.03 \\
Structure            & 12.81\\
None helpful & 5.68 \\ \bottomrule
\end{tabular}
\caption{Most useful aspects of an influence graph.}
\label{tab:ig-aspects}
\end{table}

\begin{figure}[!ht]
\centering
{\includegraphics[width=\columnwidth]{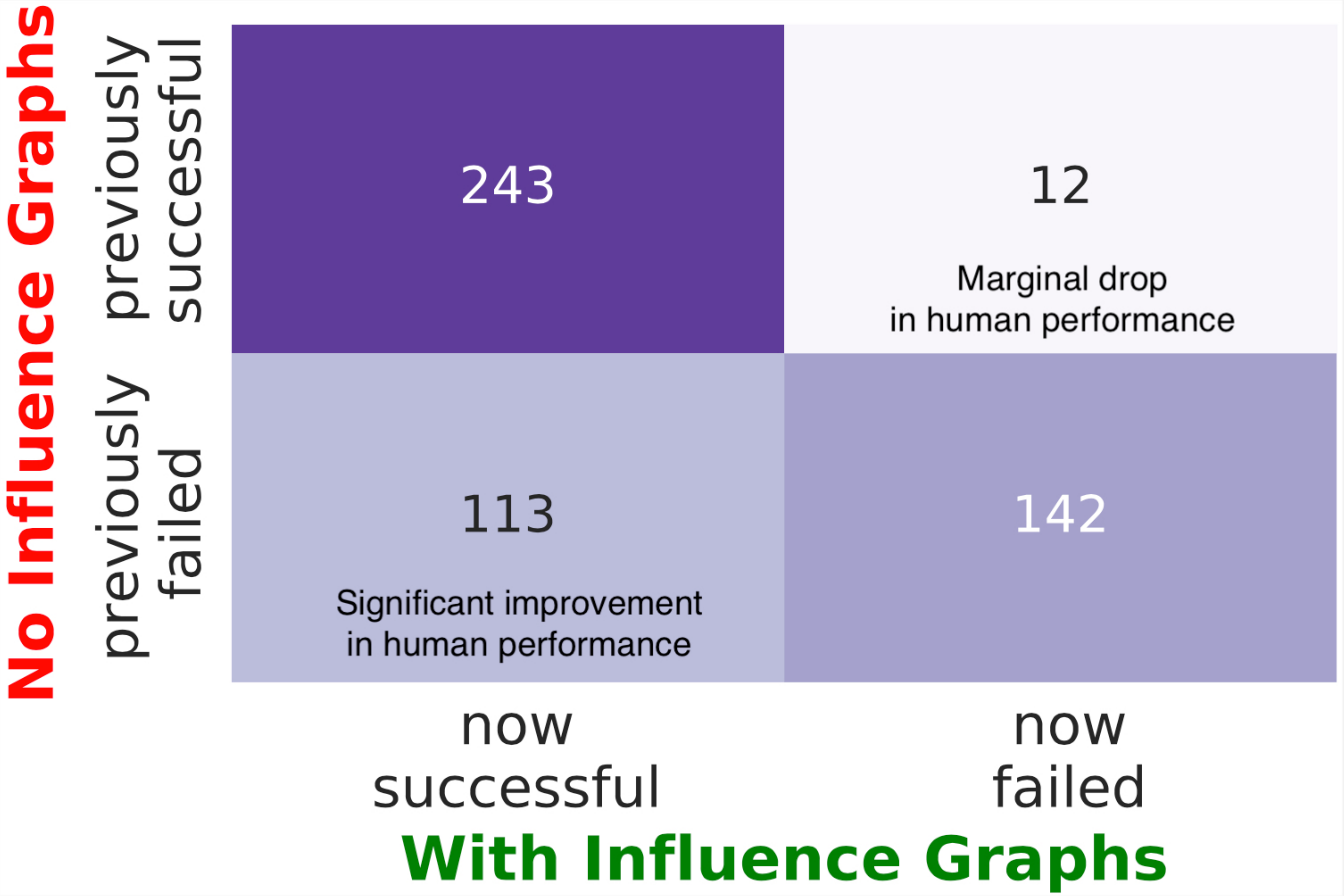}}
\caption{Human performance before and after the human judges were provided with the influence graph.}
\label{fig:cmat}
\end{figure}

\paragraph{Finding 3: Machine generated influence graphs help humans in defeasible reasoning}
Table~\ref{tab:human-results} shows that performance improves across all three tasks when the defeasible query is augmented with an influence graph.
On our challenging set of 510 queries, the overall accuracy jumps nearly 20 points from 0.50 to 0.698. 
Figure~\ref{fig:cmat} highlights that 113 queries that were previously given the wrong answers were marked correctly when augmented with the influence graphs.

\begin{table}[ht]
\centering
{
\begin{tabular}{lcc} 
\toprule
    Dataset & Human & Human     \\    
     & \cite{rudinger-etal-2020-thinking}  &  (ours)     \\    
    \toprule
    \snli &	0.461 $\pm$ 0.11 &	0.553 $\pm$ 0.11 \\
    \social	& 0.628 $\pm$ 0.07 &	0.814 $\pm$ 0.06 \\
    \atomic &0.418 $\pm$ 0.06 &	0.657 $\pm$ 0.06 \\
    \midrule 
    overall&	0.500 $\pm$ 0.04 &	\textbf{0.698 $\pm$ 0.04}\\
    \bottomrule
\end{tabular}
}
\caption{Human performance (accuracy) on the three tasks with and without generated influence graphs along with Wilson's score intervals for $\alpha = 95\%$. We tested the statistical significance of these results using the McNemar's test~\cite{mcnemar1947note} and found the results to be statistically highly significant~($p < 1e-6$).}
\label{tab:human-results}
\end{table}

%% file: sections/conclusion.tex
\section{Discussion and Conclusion} 

Our work takes the idea of using inference graphs for defeasible inference and scales up its usability by automatically generating and augmenting them to a downstream defeasible task that both humans and machines are known to find difficult.
We identify that the contextualizer and mediator nodes are crucial to defeasible inference, and show that our generated graphs generate these critical nodes effectively. 
Humans perform significantly better (20\% absolute improvement) across diverse defeasible datasets and overwhelmingly attribute their success to the mediator nodes -- giving insights into what helps and why.
In this case study, we show that machines can fill the gaps in human knowledge when for defeasible reasoning.
While we establish that humans are helped by these graphs, a further investigation
on how (and if) the graphs reinforced their beliefs, and what additional information in the graphs was beneficial to their understanding is essential. 
Furthermore, a deeper understanding of the trade-offs (time spent in answering these questions with and without the graphs) also forms important future work.

\section*{Acknowledgments}

We would like to thank Peter Clark for the thoughtful discussions, and the anonymous reviewers for valuable feedback. 

This material is partly based on research sponsored in part by the Air Force Research Laboratory under agreement number FA8750-19-2-0200. 
The U.S. Government is authorized to reproduce and distribute reprints for Governmental purposes notwithstanding any copyright notation thereon. 
The views and conclusions contained herein are those of the authors and should not be interpreted as necessarily representing the official policies or endorsements, either expressed or implied, of the Air Force Research Laboratory or the U.S. Government.
We would like to thank Google for providing the TPU machines for conducting experiments.

%% file: sections/appendix.tex
\appendix


\section{Sample input-output sequence for training \lm}
\label{sec:sampleinputoutput}
We now present a sample input-output sequence used to train out \lm for graph generation.
The input-output sample ($\V{seq_{ip}}, \V{seq_{op}}$) is presented below.
As mentioned in Section
\begin{enumerate}
    \item As described in section 2.1, each input sequence $\V{seq_{ip}}$ is formatted in a special template to be fed to the language model~(Template (1)). We show an example of the same next for a sample from our training data. {\tt \textbf{Premise: } Sunlight shines on plants. Cells with chlorophyll in them $\ldots$ other parts of the plant. \red{\textbf{ | }}\textbf{Situation} : more minerals are absorbed \red{\textbf{ | }} \textbf{Less} : LESS sugar and oxygen being produced \red{\textbf{ | }} \textbf{More} : MORE sugar and oxygen being produced}
    \item Each output graph is encoded in as a \dotf string. The output \dotf sequence $\V{seq_{op}}$ corresponding to the input shown above is:  {\tt strict digraph { "C+ : less minerals in the soil [OR] less root system" -> "S : more minerals are absorbed" [label=hurts]; "C- :more minerals in the soil [OR] a better root system" -> "S : more minerals are absorbed" [label=helps]; "S : more minerals are absorbed" -> "M- : less conversion into sugars [OR] less oxygen produced" [label=hurts]; "S : more minerals are absorbed" -> "M+ : more conversion into sugars" [label=helps]; "S- : less minerals absorbed [OR] less root system" -> "M+ : more conversion into sugars" [label=hurts]; "M- : less conversion into sugars [OR] less oxygen produced" -> "H- : LESS sugar and oxygen being produced" [label=helps]; "M- : less conversion into sugars [OR] less oxygen produced" -> "H+ : MORE sugar and oxygen being produced" [label=hurts]; "M+ : more conversion into sugars" -> "H+ : MORE sugar and oxygen being produced" [label=helps]; "M+ : more conversion into sugars" -> "H- : LESS sugar and oxygen being produced" [label=hurts]; }}
\end{enumerate}

\begin{figure*}[!ht]
\centering
{\includegraphics[scale=0.3]{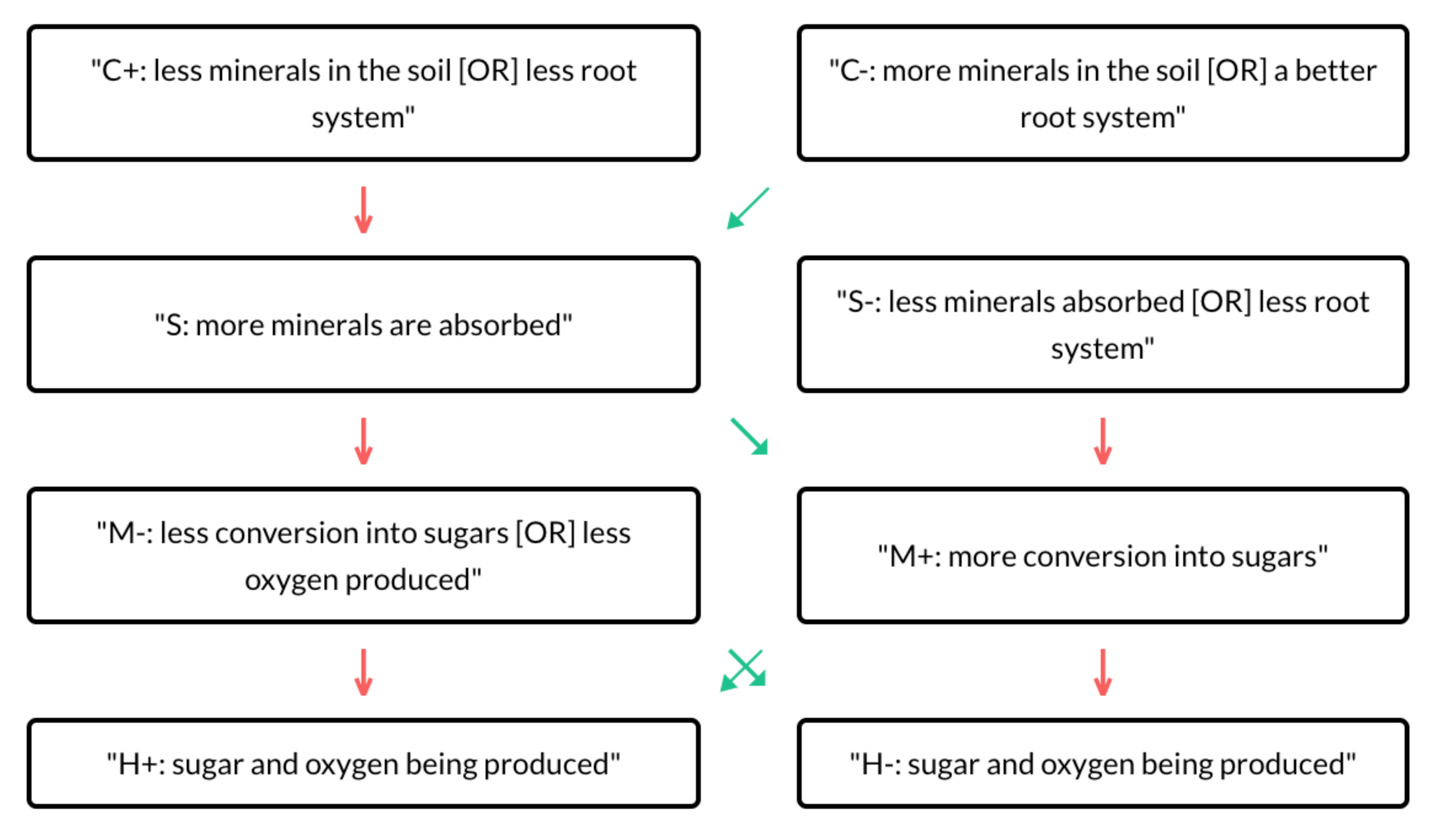}}
\caption{The influence graph corresponding to dot code shown in $\V{seq_{op}}$}
\label{fig:dotgraph}
\end{figure*}

\section{IRB Exemption}
\label{sec:irbexemption}
Our study was not an experimentation on humans (posed no identifiable risk to the human judges), did not collect any identifying information, and ensured it involved only adults. As per the IRB guidelines, this falls under the purview of human research, and we are not publishing individual workers' answers but rather the data is tallied up, much like a ``benign behavioral intervention.” This exempts us from IRB~(category 3 of Federal Regulations for Protection of Human Research Subjects https://www.hhs.gov/ohrp/regulations-and-policy/regulations/45-cfr-46/).

\section{Infrastructure and hyperparameters}
To train the T5-11B model, comprising of 11 billion parameters, we used v3-8 TPUs. The average time to train was 7 hours for about 10 epochs. We used the same hyperparameters as provided with the T5 checkpoint at \url{gs://t5-data/pretrained_models/11B}. We used maximum block size of 512 tokens, and max generation length set to 512. For decoding, we sample according to predicted distribution. We train the GPT-2 model on a Nvidia GTX 2080 Ti, and training the model takes about 30 minutes per epoch.

We use the medium (355M) variant of \textsc{gpt-2}~\cite{radford2019language} with 24 layers, 1024 hidden size, 16 attention heads.

\section{Details of our Mechanical Turk Setup}
\begin{figure}[!h]
\centering
\includegraphics[width=0.4\columnwidth]{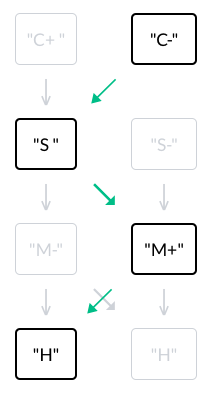}
\caption{Part of the generated influence graph that is presented in the hit.}
\label{appendix:mturk_ig_skeleton}
\end{figure}

We follow the same instructions for humans as \cite{rudinger-etal-2020-thinking}\footnote{We are grateful to the authors of \cite{rudinger-etal-2020-thinking} for sharing their mechanical turk setup template with us.}, and only additionally provided instructions for the inference graph. We used a pool of 230 annotators that were previously qualified and selected to do the defeasible inference task, thus providing a fair comparison to their setup. Eventually 12 workers out of these 230 workers worked on our HITs. The graph we showed to humans was a subgraph of the inference graph, where the selected path has the relevant content from the inference graph to avoid showing redundant opposite edges. These redundant edges are useful in training a model as the model must jointly predict all the nodes, but this is redundant for humans. Figure \ref{appendix:mturk_ig_skeleton} shows this subgraph. 

\subsection{A sample HIT}
\label{sec:samplehiit}
\begin{figure*}[!ht]
\centering
\includegraphics[width=\textwidth]{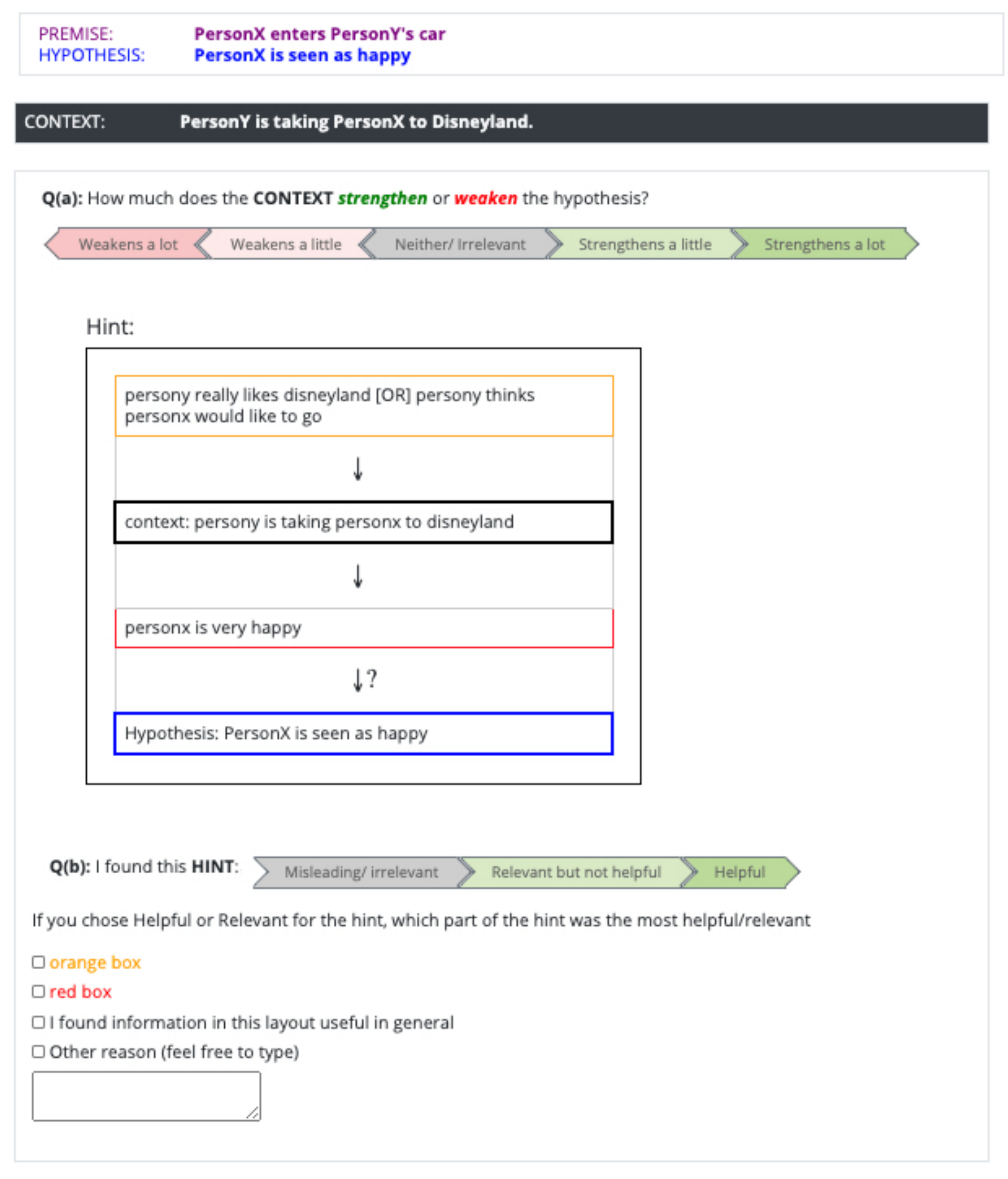}
\caption{A sample HIT in mechanical turk.}
\label{appendix:mturk}
\end{figure*}

We now show a sample HIT in Figure \ref{appendix:mturk}. We had two set of annotations in every HIT.

\subsection{Examples that helped humans}
\begin{figure}[!t]
\centering
\includegraphics[width=1.0\columnwidth]{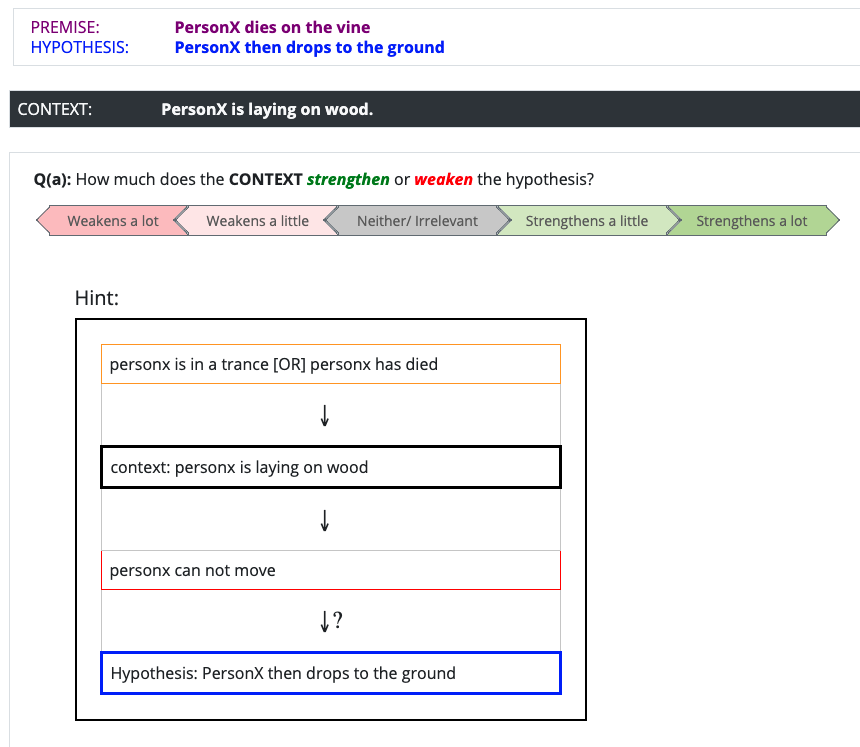}
\caption{An example where the graph helped the human in getting the correct answer, that humans were unsuccessful on, in the past.}
\label{appendix:mturk_ig_prev_wrong_now_right1}
\end{figure}

\begin{figure}[!b]
\centering
\includegraphics[width=1.0\columnwidth]{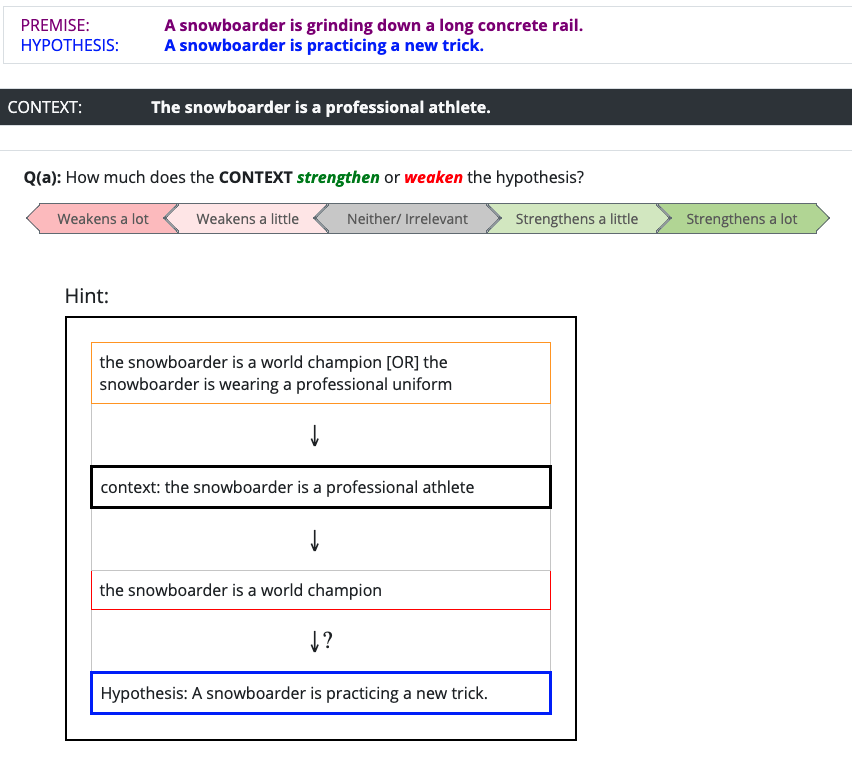}
\caption{Another example where the graph helped the human in getting the correct answer, that humans were unsuccessful on, in the past.}
\label{appendix:mturk_ig_prev_wrong_now_right2}
\end{figure}

Next, we show two examples (Figure \ref{appendix:mturk_ig_prev_wrong_now_right1}, Figure \ref{appendix:mturk_ig_prev_wrong_now_right2}) where humans were previously unsuccessful on this answer (in the original setup of \cite{rudinger-etal-2020-thinking}), and were successful now having looked at the inference graphs. The humans marked that the mediator nodes and the contextualizer nodes provide useful information.

\section{Dataset}
\label{sec:datasetdetails}
\begin{table}[!h]
 \resizebox{\columnwidth}{!}{
 \small
\begin{tabular}{llrr}
\toprule
Dataset & Split & \# Samples & Total\\ \midrule
 \multirow{3}{*}{\wiqa} & train & 1522 & \multirow{3}{*}{2107} \\
      & test   & 189  \\
& dev  & 152 \\ \midrule
 \multirow{3}{*}{\atomic} & train & 35,001 & \multirow{3}{*}{42,977} \\
      & test   & 4137  \\
& dev  & 3839 \\ \midrule
 \multirow{3}{*}{\social} & train & 88,675 & \multirow{3}{*}{92,295} \\
      & test   & 1836  \\
& dev  & 1784 \\ \midrule
 \multirow{3}{*}{\snli} & train & 77,015 & \multirow{3}{*}{95,795} \\
      & test   & 9438  \\
& dev  & 9342 \\
\bottomrule
\end{tabular}
}
\caption{Number of samples in each dataset by split. \atomic, \snli, \social are available at~\url{https://github.com/rudinger/defeasible-nli}, \wiqa is avilable at~\url{https://allenai.org/data/wiqa}}
\label{table:data-split}

\end{table}